\documentclass[letterpaper]{article}
\usepackage{natbib,alifexi}

\usepackage{graphicx}
\usepackage{amssymb}
\usepackage{times}
\usepackage{verbatim}
\usepackage{amsmath}
\usepackage{algorithmic,algorithm}

\DeclareGraphicsExtensions{.eps}
\graphicspath{{figures/}}

\title{The Connectivity of NK Landscapes' Basins: A Network Analysis}
\author{S{\'e}bastien  Verel$^{1}$, Gabriela Ochoa$^{2}$ \and Marco Tomassini$^{3}$  \\
\mbox{}\\
$^1$University of Nice Sophia-Antipolis / CNRS, Nice, France\\
$^2$University of Nottingham, Nottingham,UK\\
$^3$University of Lausanne, Lausanne, Switzerland \\
{\tt verel@i3s.unice.fr, gxo@cs.nott.ac.uk,
Marco.Tomassini@unil.ch}}
\begin{document}
\maketitle

\begin{abstract}
We propose a network characterization of combinatorial fitness
landscapes by adapting the notion of {\em inherent networks}
proposed for energy surfaces~\cite{doye02}. We use the well-known
family of $NK$ landscapes as an example. In our case the inherent
network is the graph where the vertices represent the local maxima
in the landscape, and the edges account for the transition
probabilities between their corresponding basins of attraction. We
exhaustively extracted such networks on representative small $NK$
landscape instances, and performed a statistical characterization of
their properties. We found that most of these network properties can
be related to the search difficulty on the underlying $NK$
landscapes with varying values of $K$.

\end{abstract}

\section{Introduction}

Local optima are the very feature of a landscape that makes it
rugged. Therefore, an understanding of the distribution of local
optima is of utmost importance for the understanding of a landscape.
Combinatorial landscapes refer to the finite search spaces generated by important discrete
problems such as the traveling salesman problem and many others.
A property of some combinatorial landscapes, which has been often
observed, is that on average, local optima
are  much closer to the optimum than are randomly chosen points,
and closer to each other than random points would be. In other
words, the local optima are not randomly distributed, rather they
tend to be clustered in a "central massif" (or "big valley" if we
are minimising). This globally convex landscape structure has been
observed in the $NK$ family of landscapes \cite{kauffman93}, and in
other combinatorial optimization problems, such as the traveling
salesman problem \cite{boese94}, graph bipartitioning
\cite{merz98}, and flowshop scheduling \cite{reeves99}.

In this study we seek to provide fundamental new insights into the
structural organization of the local optima in $NK$
landscapes, particularly into the connectivity of their basins of
attraction. Combinatorial landscapes can be seen as a graph whose
vertices are the possible configurations. If two configurations can
be transformed into each other by a suitable operator move, then we
can trace an edge between them. The resulting graph, with an
indication of the fitness at each vertex, is a representation of the
given problem fitness landscape. A useful  simplification of the
graphs for the energy landscapes of atomic clusters was introduced
in \cite{doye02,doye05}. The idea consists in taking as vertices of
the graph not all the possible configurations, but only those that
correspond to energy minima. For atomic clusters these are
well-known, at least for relatively small assemblages. Two minima
are considered connected, and thus an edge is traced between them,
if the energy barrier separating them is sufficiently low. In this
case there is a transition state, meaning that the system can jump
from one minimum to the other by thermal fluctuations going through
a saddle point in the energy hyper-surface. The values of these
activation energies are mostly known experimentally or can be
determined by simulation. In this way, a network can be built which
is called the ``inherent structure'' or ``inherent network''
in~\cite{doye02}.\\
We propose a network characterization of combinatorial fitness
landscapes by adapting the notion of {\em inherent networks}
described above. We use the well-known family of $NK$ landscapes as
an example because they are a useful tunable benchmark that can provide
interesting information for more realistic combinatorial landscapes.
 In our case the inherent network is the graph where the
vertices are all the local maxima and the edges account for
transition probabilities between their corresponding basins of
attraction. We exhaustively extract such networks on representative
small $NK$ landscape instances, and perform a statistical
characterization of their properties. Our analysis was inspired, in
particular, by the work of \cite{doye02,doye05} on energy
landscapes, and in general, by the field of complex
networks~\cite{newman03}. The study of networks
has exploded across the academic world since the
late 90's. Researchers from the mathematical, biological, and social
sciences have made substantial progress on some previously
intractable problems, bringing new techniques, reformulating old
ideas, and uncovering unexpected connections between seemingly
different problems. We aim here at bringing the tools of network
analysis for the study of problem hardness in combinatorial
optimization.\\
The next section describes how combinatorial landscapes are mapped
onto networks, and includes the relevant definitions and algorithms
used in our study. The empirical network analysis of our selected
$NK$ landscape instances is presented next, followed by our
conclusions and ideas for future work.

\section{Landscapes as Networks}
\label{landscapes}

To model a physical energy landscape as a network, \cite{doye05}
needed to decide first on a definition both of a state of the system
and how two states were connected. The states and their connections
will then provide the nodes and edges of the network. For systems
with continuous degrees of freedom, the author achieved this through
the `inherent structure' mapping. In this mapping each point in
configuration space is associated with the minimum (or `inherent
structure') reached by following a steepest-descent path from that
point. This mapping divides the configuration space into basins of
attraction surrounding each minimum on the energy landscape.

Our goal is to adapt this idea to the context of combinatorial
optimization. In our case, the nodes of the graph can be
straightforwardly defined as the local maxima of the landscape.
These maxima are obtained exhaustively by running a best-improvement
local search algorithm ({\em HillClimbing}, see Algorithm 1) from
every configuration of the search space. The definition of the
edges, however, is a much more delicate matter. In our initial
attempt~\cite{gecco08} we considered that two maxima $i$ and $j$
were connected (with an undirected and unweighted edge), if there
exists at least one pair of  solutions at Hamming distance one $s_i$
and $s_j$, one in each basin of attraction ($b_i$ and $b_j$). We
found empirically on small instances of $NK$ landscapes, that such
definition produced densely connected graphs, with very low ($\leq
2$) average path length between nodes for all $K$. Therefore, apart
from the already known increase in the number of optima with
increasing $K$, no other network property accounted for the increase
in search difficulty. Furthermore, a single pair of neighbors
between adjacent basins, may not realistically account for actual
basin transitions occurring when using common heuristic search
algorithms. These considerations, motivated us to search for an
alternative definition of the edges connecting local optima. In
particular, we decided to associate weights to the edges that
account for the transition probabilities between the basins of
attraction of the local optima. More details on the relevant
algorithms and formal definitions are given below.

\subsection{Definitions and Algorithms}
\label{sec:defs}

\textbf{Definition:} Fitness landscape.\\
A landscape is a triplet $(S, V, f)$ where $S$ is a set of potential
solutions i.e. a search space, $V : S \longrightarrow 2^S$, a
neighborhood structure, is a function that assigns to every $s \in
S$ a set of neighbors $V(s)$, and $f : S \longrightarrow R$ is a
fitness function that can be pictured as the \textit{height} of the
corresponding solutions.

In our study, the search space is composed by binary strings of
length $N$, therefore its size is $2^N$. The neighborhood is defined
by the minimum possible move on a binary search space, that is, the
1-move or bit-flip operation. In consequence, for any given string
$s$ of length $N$, the neighborhood size is $|V(s)| = N$.

The $HillClimbing$ algorithm to determine the local optima and
therefore define the basins of attraction, is given below:

\begin{algorithm}
\caption{{\em HillClimbing}} \label{algoHC}
\begin{algorithmic}
\STATE Choose initial solution $s \in S$ \REPEAT
    \STATE choose $s^{'} \in V(s)$ such that $f(s^{'}) = max_{x \in V(s)}\ f(x)$
        \IF{$f(s) < f(s^{'})$}
            \STATE $s \leftarrow s^{'}$
    \ENDIF
\UNTIL{$s$ is a  Local optimum}
\end{algorithmic}
\end{algorithm}

\textbf{Definition:} Local optimum.\\
A local optimum is a solution $s^{*}$ such that $\forall  s \in
V(s^{*})$, $f(s) < f(s^{*})$.

\noindent The $HillClimbing$ algorithm defines a mapping from the
search space $S$ to the set of locally optimal solutions $S^*$.

\textbf{Definition:} Basin of attraction.\\
The basin of attraction of a local optimum $i \in S$ is the set $b_i = \{
s \in S ~|~ HillClimbing(s) = i \}$. The size of the basin of
attraction of a local optima $i$ is the cardinality of $b_i$.

 \textbf{Definition:} Edge weight.\\
Notice that for a non-neutral fitness landscapes, as are $NK$
landscapes, the basins of attraction as defined above, produce a
partition of the configuration space $S$. Therefore, $S = \cup_{i
\in S^{*}} b_i$ and $\forall i \in S$ $\forall j \not= i$, $b_i \cap
b_j = \emptyset$

\noindent For each solutions $s$ and $s^{'}$, let us define $p(s
\rightarrow s^{'} )$ as the probability
to pass from $s$ to $s^{'}$ with the bit-flip operator. In the case
of binary strings of size $N$, and the neighborhood defined by the
bit-flip operation, there are $N$ neighbors for each solution,
therefore:

\noindent if $s^{'} \in V(s)$ , $p(s \rightarrow s^{'} ) = \frac{1}{N}$ and \\
if $s^{'} \not\in V(s)$ , $p(s \rightarrow s^{'} ) = 0$.

\noindent We can now define the probability to pass from a solution $s
\in S$ to a solution belonging to the basin $b_j$, as:
$$
p(s \rightarrow b_j ) = \sum_{s^{'} \in b_j} p(s \rightarrow s^{'} )
$$

\noindent Notice that $p(s \rightarrow b_j ) \leq 1$.

\noindent Thus, the total probability of going from basin $b_i$ to
basin $b_j$ is the average over all $s \in b_i$ of the transition
probabilities  to solutions $s^{'} \in b_j$ :

$$p(b_i \rightarrow b_j) = \frac{1}{\sharp b_i} \sum_{s \in b_i} p(s \rightarrow b_j )$$

\noindent $\sharp b_i$ is the size of the basin $b_i$.
We are now prepared to define our `inherent' network or network of
local optima.

\textbf{Definition:} Local optima network.\\
The local optima network $G=(S^*,E)$ is the graph where the nodes
are the local optima \footnote{Since each maximum has its associated
basin, $G$ also describes the interconnection of basins.}, and there
is an edge $e_{ij} \in E$ with the weight $w_{ij} = p(b_i
\rightarrow b_j)$ between two nodes $i$ and $j$ if $p(b_i
\rightarrow b_j) > 0$.

\noindent According to our definition of weights, $w_{ij} = p(b_i
\rightarrow b_j)$ may be different than $w_{ji} = p(b_j \rightarrow
b_i)$. Two weights are needed in general, and we have an
oriented transition graph.

\section{Empirical Basin and  Network Analysis}
\label{analysis}

The $NK$ family of landscapes \cite{kauffman93} is a
problem-independent model for constructing multimodal landscapes
that can gradually be tuned from smooth to rugged. In the model, $N$
refers to the number of (binary) genes in the genotype (i.e. the
string length) and $K$ to the number of genes that influence a
particular gene (the epistatic interactions). By increasing the
value of $K$ from 0 to $N-1$, $NK$ landscapes can be tuned from
smooth to rugged. The $k$ variables that form the context of the
fitness contribution of gene $s_i$ can be chosen according to
different models. The two most widely studied models are the {\em
random neighborhood} model, where the $k$  variables are chosen
randomly according to a uniform distribution among the $n-1$
variables other than $s_i$, and the {\em adjacent neighborhood}
model, in which the $k$ variables that are closest to $s_i$ in a
total ordering $s_1, s_2, \ldots, s_n$ (using periodic boundaries).
No significant differences between the two models were found in
\cite{kauffman93} in terms of  global properties of the respective
families of landscapes, such as mean number of local optima or
autocorrelation length. Similarly, our preliminary studies on the
characteristics of the $NK$ landscape optima networks did not show
noticeable differences between the two neighborhood models.
Therefore, we conducted our full study on the more general random
model.

In order to avoid sampling problems that could bias the results, we
used the largest values of $N$ that can still be analysed
exhaustively with reasonable computational resources. We thus
extracted the local optima networks of landscape instances with $N =
{14, 16, 18}$, and $K = {2,4,6,..., N-2,N-1}$. For each pair of $N$
and $K$ values, 30 randomly generated instances were explored. Therefore, the networks
statistics reported below represent the average behaviour of 30
independent instances.

\enlargethispage{\baselineskip}

\subsection{Basins of Attraction}

Besides the maxima network, it is useful to describe the associated
basins of attraction as these play a key role in search algorithms.
Furthermore, some characteristics of the basins can be related to
the optima network features. The notion of the basin of attraction
of a local maximum has been presented before. We
have exhaustively computed the size and number of all the basins of
attraction for $N=16$ and $N=18$ and for all even $K$ values plus
$K=N-1$. In this section, we analyze the basins of attraction from
several points of view as it is described below.

\subsubsection{Global optimum basin size versus $\bf K$.}

\begin{figure} [!ht]
\begin{center}
\includegraphics[width=6.3cm] {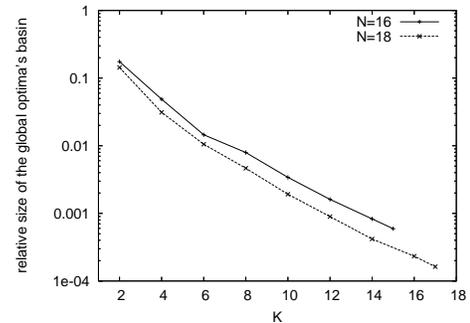} \protect \\
\caption{Average of the relative size of the basin corresponding to
the global maximum for each K over 30 landscapes.\label{nbasins}}
\end{center}
\end{figure}

In Fig.~\ref{nbasins} we plot the average size of the basin
corresponding to the global maximum for $N=16$ and $N=18$, and all
values of $K$ studied. The trend is clear: the basin shrinks very
quickly with increasing $K$. This confirms that the higher the $K$
value, the more difficult for an stochastic search algorithm to
locate the basin of attraction of the global optimum

\subsubsection{Number of basins of a given size.}

\begin{figure} [!ht]
\begin{center}
    \mbox{\includegraphics[width=6.3cm]{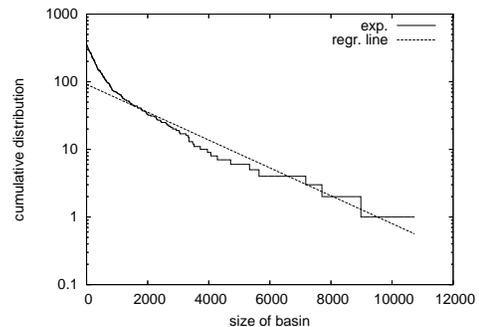} } \protect \\
\caption{Cumulative distribution of the number of basins of a given
size with regression line. A representative landscape with $N=18$, $K=4$ is
visualized. A lin-log scale is used. \label{bas-18-size}}
\end{center}
\end{figure}

\begin{table}[!ht]
\begin{center}
\small \caption{Correlation coefficient ($\bar{\rho}$), and linear
regression coefficients (intercept ($\bar{\alpha})$ and slope
($\bar{\beta}$)) of the relationship between the basin size of
optima and the cumulative number of nodes of a given (basin) size (
in logarithmic scale: $\log(p(s)) = \alpha + \beta s + \epsilon$).
The average and standard deviation values over 30 instances, are
shown.}
\label{tab:cumulSize} \vspace{0.2cm}
\begin{tabular}{|c|c|c|c|}
\hline
\multicolumn{4}{|c|}{$N = 16$} \\
\hline
 $K$ & $\bar \rho$ & $\bar \alpha$ & $\bar{\beta}$ \\
\hline
2   & $-0.944_{0.0454}$ & $2.89_{0.673}$ & $-0.0003_{0.0002}$  \\
4   & $-0.959_{0.0310}$ & $4.19_{0.554}$ & $-0.0014_{0.0006}$  \\
6   & $-0.967_{0.0280}$ & $5.09_{0.504}$ & $-0.0036_{0.0010}$  \\
8   & $-0.982_{0.0116}$ & $5.97_{0.321}$ & $-0.0080_{0.0013}$  \\
10  & $-0.985_{0.0161}$ & $6.74_{0.392}$ & $-0.0163_{0.0025}$  \\
12  & $-0.990_{0.0088}$ & $7.47_{0.346}$ & $-0.0304_{0.0042}$  \\
14  & $-0.994_{0.0059}$ & $8.08_{0.241}$ & $-0.0508_{0.0048}$  \\
15  & $-0.995_{0.0044}$ & $8.37_{0.240}$ & $-0.0635_{0.0058}$  \\
\hline \hline
\multicolumn{4}{|c|}{$N = 18$} \\
\hline
2   & $-0.959_{0.0257}$ & $3.18_{0.696}$ & $-0.0001_{0.0001}$  \\
4   & $-0.960_{0.0409}$ & $4.57_{0.617}$ & $-0.0005_{0.0002}$  \\
6   & $-0.967_{0.0283}$ & $5.50_{0.520}$ & $-0.0015_{0.0004}$  \\
8   & $-0.977_{0.0238}$ & $6.44_{0.485}$ & $-0.0037_{0.0007}$  \\
10  & $-0.985_{0.0141}$ & $7.24_{0.372}$ & $-0.0077_{0.0011}$  \\
12  & $-0.989_{0.0129}$ & $7.98_{0.370}$ & $-0.0150_{0.0019}$  \\
14  & $-0.993_{0.0072}$ & $8.69_{0.276}$ & $-0.0272_{0.0024}$  \\
16  & $-0.995_{0.0056}$ & $9.33_{0.249}$ & $-0.0450_{0.0036}$  \\
17  & $-0.992_{0.0113}$ & $9.49_{0.386}$ & $-0.0544_{0.0058}$  \\
\hline
\end{tabular}
\end{center}
\end{table}

Fig.~\ref{bas-18-size} shows the cumulative distribution of the
number of basins of a given size (with regression line) for a
representative instances with $N=18$, $K = 4$ . Table
~\ref{tab:cumulSize} shows the average (of 30 independent
landscapes) correlation coefficients and linear regression
coefficients (intercept ($\bar{\alpha})$ and slope ($\bar{\beta}$))
between the number of nodes and the basin sizes for
instances with $N=16, 18$.
Notice that
distribution decays exponentially or faster for the lower $K$ and it
is closer to exponential for the higher $K$. This could be
relevant to theoretical studies that estimate the size of attraction
basins (see for example \cite{garnier01}). These studies often
assume that the basin sizes are uniformly distributed, which is not
the case for the $NK$ landscapes studied here. From the slopes $\bar
\beta$ of the regression lines (table~\ref{tab:cumulSize}) one can
see that high values of $K$ give rise to steeper distributions
(higher $\bar \beta$ values). This indicates that there are fewer
basins of large size for large values of $K$. Basins
are thus broader for low values of $K$, which is consistent with the fact
that those landscapes are smoother.

\subsubsection{Fitness of local optima versus their basin sizes.}
\begin{figure} [!ht]
\begin{center}
    \mbox{\includegraphics[width=6.3cm]{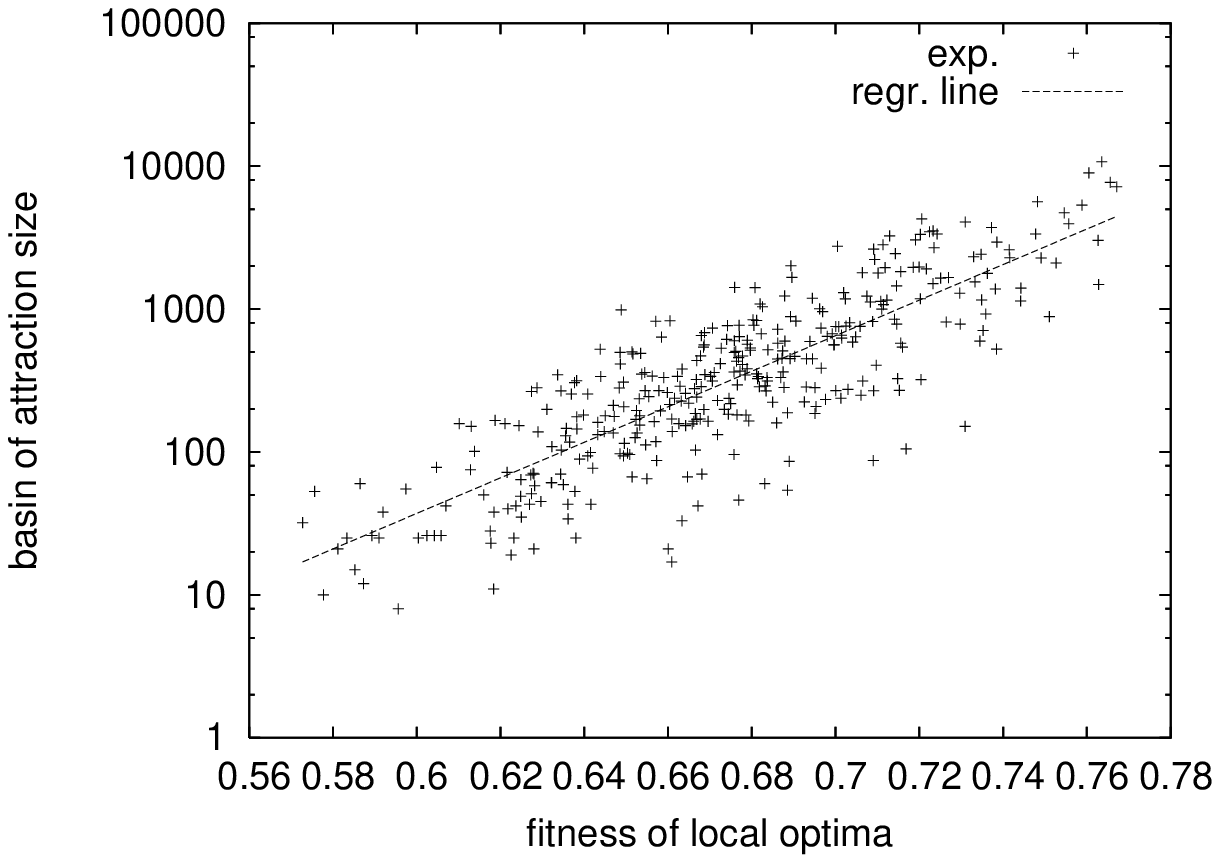} } \protect \\
    \mbox{\includegraphics[width=6.3cm]{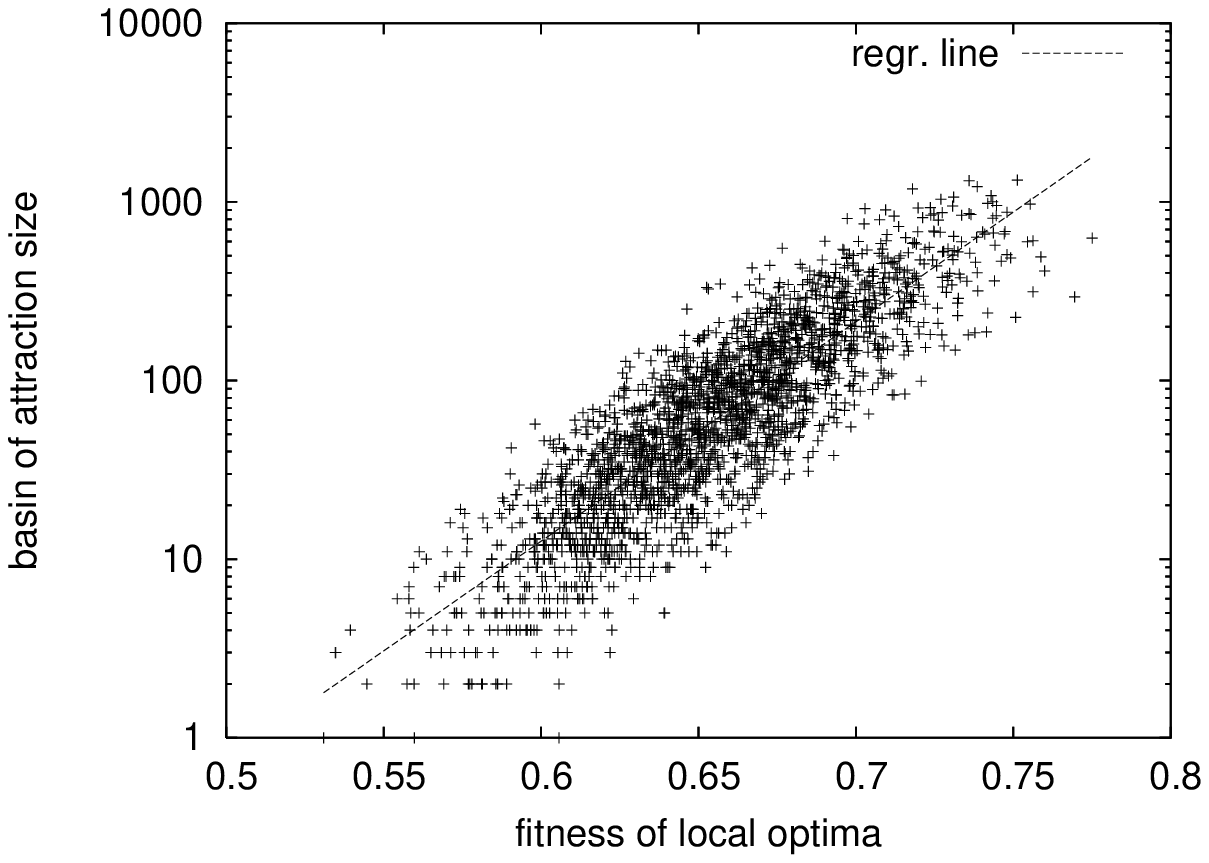} } \protect \\
\caption{Correlation between the fitness of local optima and their
corresponding basin sizes, for two representative instances with
$N=18$, $K=4$ (top) and $K=8$ (bottom). \label{fig:cor_fit-size}}
\end{center}
\end{figure}
The scatter-plots in Fig.~\ref{fig:cor_fit-size} illustrate the
correlation between the basin sizes of local maxima (in logarithmic
scale) and their fitness values. Two representative instances for
$N$ = 18 and $K$ = 4, 8 are shown. Notice that there is a clear
positive correlation between the fitness values of maxima and their
basins' sizes. In other words, the higher the peak the wider tend to
be its basin of attraction. Therefore, on average, with a stochastic
local search algorithm, the global optimum would be easier to find
than any other local optimum. This may seem surprising. But, we have
to keep in mind  that as the number of local optima increases (with
increasing $K$), the global optimum basin is more difficult to reach
by a stochastic local search algorithm (see Fig.~\ref{nbasins}).
This observation offers a mental picture of $NK$ landscapes: we can
consider the landscape as composed of a large number of mountains
(each corresponding to a basin of attraction), and those mountains
are wider the taller the hilltops. Moreover, the size of a mountain
basin grows exponentially with its hight.

\section{General Network Statistics}
\label{net-stat}

We now briefly describe the  statistical measures used for our
analysis of maxima networks.

The standard  clustering coefficient \cite{newman03} does not
consider weighted edges. We thus use the {\em weighted clustering}
measure proposed by \cite{bart05}, which combines the topological
information with the weight distribution of the network:

$$c^{w}(i) = \frac{1}{s_i(k_i - 1)} \sum_{j,h} \frac{w_{ij} + w_{ih}}{2} a_{ij} a_{jh} a_{hi}$$
where $s_i = \sum_{j \not= i} w_{ij}$, $a_{nm} = 1$ if $w_{nm} > 0$,
$a_{nm} = 0$ if $w_{nm} = 0$ and $k_i = \sum_{j \not= i} a_{ij}$.

For each triple formed in the neighborhood of the vertex $i$,
$c^{w}(i)$ counts the weight of the two participating edges of the
vertex $i$. $C^w$ is defined as the weighted clustering coefficient
averaged over all vertices of the network.

The standard topological characterization of networks is obtained by
the analysis of the probability distribution $p(k)$ that a randomly
chosen vertex has degree $k$. For our weighted networks, a
characterization of weights is obtained by the {\em connectivity and
weight distribution} $p(w)$ that any given edge has weight $w$.

In our study, for each node $i$, the sum of weights from the node
$i$ is equal to $1$. So, an important measure is the weight $w_{ii}$
of self-connecting edges (remaining in the same node). We have the
relation: $ w_{ii} + s_i = 1$. The vertex {\em strength}, $s_i$, is
defined as $s_i = \sum_{j \in {\cal V}(i) - \{i\}}  w_{ij}  $, where
the sum is over the set ${\cal V}(i) - \{i\}$ of neighbors of
$i$~\cite{bart05}. The strength of a node is a generalization of
the node's connectivity giving information about the number and
importance of the edges.

Another network measure we report here is {\em disparity}
\cite{bart05} $Y_{2}(i)$, which measures how heterogeneous is the
contributions of the edges of node $i$ to the total weight
(strength):

$$Y_{2}(i) = \sum_{j \not= i} \left( \frac{w_{ij}}{s_i} \right)^2 $$

The disparity could be averaged over the node with the same degree
$k$. If all weights are nearby of $s_i/k$, the disparity for nodes
of degree $k$ is nearby $1/k$.

Finally, in order to compute the average distance (shortest path)
between two nodes on the optima network of a given landscape, we
considered the expected number of bit-flip mutations to pass from
one basin to the other. This expected number can be computed by
considering the inverse of the transition probabilities between
basins. In other words, if we attach to the edges the inverse of the
transition probabilities, this value would represent the average
number of random mutations to pass from one basin to the other. More
formally, the distance (expected number of bit-flip mutations)
between two nodes is defined by $d_{ij} = 1 / w_{ij}$ where $w_{ij}
= p(b_i \rightarrow b_j)$. Now, we can define the length of a path
between two nodes as being the sum of these distances along the
edges that connect the respective basins.

\enlargethispage{\baselineskip}

\begin{table*}[!ht]
\begin{center}
\small \caption{$NK$ landscapes network properties.  Values are
averages over 30 random instances, standard deviations are shown as
subscripts. $n_v$ and $n_e$ represent the number of vertexes and
edges (rounded to the next integer), $\bar C^{w}$, the mean weighted
clustering coefficient. $\bar Y$ represent the mean disparity
coefficient, $\bar d$ the mean path length (see text for
definitions). } \label{tab:statistics} \vspace{0.2cm}
\begin{tabular}{|c|c|c|c|c|c|}
\hline
$K$ & $\bar n_v$ & $\bar n_e$ & $\bar C^{w}$ & $\bar Y$ & $\bar d$ \\
\hline
\multicolumn{6}{|c|}{$N = 14$} \\
\hline
2  &    $14_{ 6}$  & $200_{131}$      & $0.98_{0.0153}$ & $0.367_{0.0934}$ & $76_{194}$ \\
4  &    $70_{10}$  & $3163_{766}$     & $0.92_{0.0139}$ & $0.148_{0.0101}$ & $89_{6}$   \\
6  &   $184_{15}$  & $12327_{1238}$   & $0.79_{0.0149}$ & $0.093_{0.0031}$ & $119_{3}$  \\
8  &   $350_{22}$  & $25828_{1801}$   & $0.66_{0.0153}$ & $0.070_{0.0020}$ & $133_{2}$  \\
10 &   $585_{22}$  & $41686_{1488}$   & $0.54_{0.0091}$ & $0.058_{0.0010}$ & $139_{1}$  \\
12 &   $896_{22}$  & $57420_{1012}$   & $0.46_{0.0048}$ & $0.052_{0.0006}$ & $140_{1}$  \\
13 &  $1085_{20}$  & $65287_{955}$    & $0.42_{0.0045}$ & $0.050_{0.0006}$ & $139_{1}$  \\

\hline \hline
\multicolumn{6}{|c|}{$N = 16$} \\
\hline
2  &     $33_{15}$ & $516_{358}$      & $0.96_{0.0245}$ & $0.326_{0.0579}$ & $56_{14}$   \\
4  &    $178_{33}$ & $9129_{2930}$    & $0.92_{0.0171}$ & $0.137_{0.0111}$ & $126_{8}$   \\
6  &    $460_{29}$ & $41791_{4690}$   & $0.79_{0.0154}$ & $0.084_{0.0028}$ & $170_{3}$   \\
8  &    $890_{33}$ & $93384_{4394}$   & $0.65_{0.0102}$ & $0.062_{0.0011}$ & $194_{2}$   \\
10 &  $1,470_{34}$ & $162139_{4592}$  & $0.53_{0.0070}$ & $0.050_{0.0006}$ & $206_{1}$   \\
12 &  $2,254_{32}$ & $227912_{2670}$  & $0.44_{0.0031}$ & $0.043_{0.0003}$ & $207_{1}$   \\
14 &  $3,264_{29}$ & $290732_{2056}$  & $0.38_{0.0022}$ & $0.040_{0.0003}$ & $203_{1}$   \\
15 &  $3,868_{33}$ & $321203_{2061}$  & $0.35_{0.0022}$ & $0.039_{0.0004}$ & $200_{1}$   \\

\hline \hline
\multicolumn{6}{|c|}{$N = 18$} \\
\hline
2  &     $50_{25}$ & $1579_{1854}$    & $0.95_{0.0291}$ & $0.307_{0.0630}$ & $73_{15}$   \\
4  &    $330_{72}$ & $26266_{7056}$   & $0.92_{0.0137}$ & $0.127_{0.0081}$ & $174_{9}$   \\
6  &    $994_{73}$ & $146441_{18685}$ & $0.78_{0.0155}$ & $0.076_{0.0044}$ & $237_{5}$   \\
8  &  $2,093_{70}$ & $354009_{18722}$ & $0.64_{0.0097}$ & $0.056_{0.0012}$ & $273_{2}$   \\
10 &  $3,619_{61}$ & $620521_{20318}$ & $0.52_{0.0071}$ & $0.044_{0.0007}$ & $292_{1}$   \\
12 &  $5,657_{59}$ & $899742_{14011}$ & $0.43_{0.0037}$ & $0.038_{0.0003}$ & $297_{1}$   \\
14 &  $8,352_{60}$ & $1163640_{11935}$& $0.36_{0.0023}$ & $0.034_{0.0002}$ & $293_{1}$   \\
16 & $11,797_{63}$ & $1406870_{6622}$ & $0.32_{0.0012}$ & $0.032_{0.0001}$ & $283_{1}$   \\
17 & $13,795_{77}$ & $1524730_{4818}$ & $0.30_{0.0009}$ & $0.032_{0.0001}$ & $277_{1}$   \\
\hline
\end{tabular}
\end{center}
\end{table*}

\subsection{Detailed Study of Network Features}

In this section we study in more depth some network features which
can be related to stochastic local search difficulty on the
underlying fitness landscapes. Table \ref{tab:statistics} reports
the average (over 30 independent instances for each $N$ and $K$) of
the network properties described. $\bar n_v$ and $\bar n_e$ are,
respectively, the mean number of vertices and the mean number of
edges of the graph for a given $K$ rounded to the next integer.
$\bar C^{w}$ is the mean weighted clustering coefficient. $\bar Y$
is the mean disparity, and $\bar d$ is the mean path length.

\subsubsection{Clustering Coefficients.}

The fourth column of table \ref{tab:statistics} lists the average
values of the weighted clustering coefficients for all $N$ and $K$.
It is apparent that the clustering coefficients decrease regularly
with increasing $K$ for all $N$. For the standard unweighed
clustering, this would mean that the larger $K$ is, the less likely
that two maxima which are connected to a third one are themselves
connected. Taking weights, i.e.~transition probabilities into
account this means that either there are fewer transitions between
neighboring basins for high $K$, and/or the transitions are less
likely to occur. This confirms from a network point of view the
common knowledge that search difficulty increases with $K$.

\subsubsection{Shortest Path to the Global Optimum.}

\begin{figure} [!ht]
\begin{center}
    \mbox{\includegraphics[width=6.3cm]{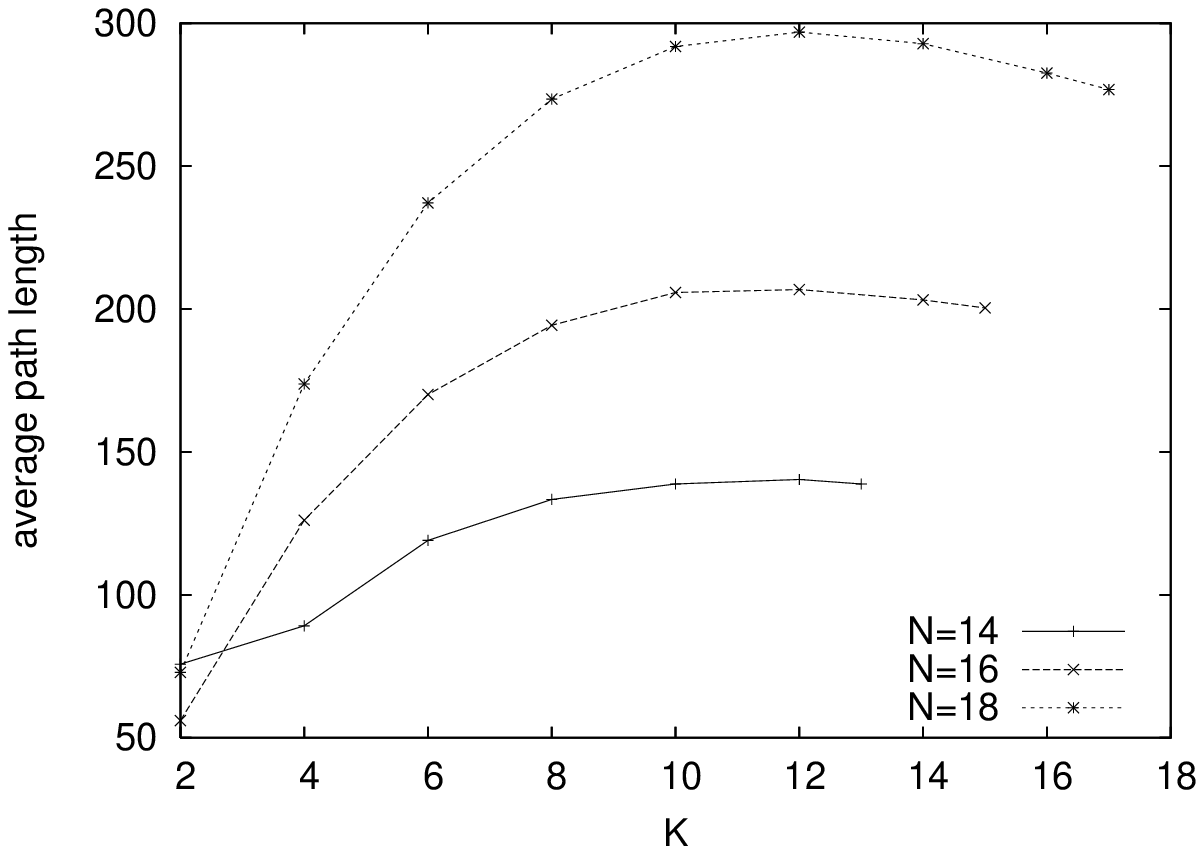} } \protect \\
    \mbox{\includegraphics[width=6.3cm]{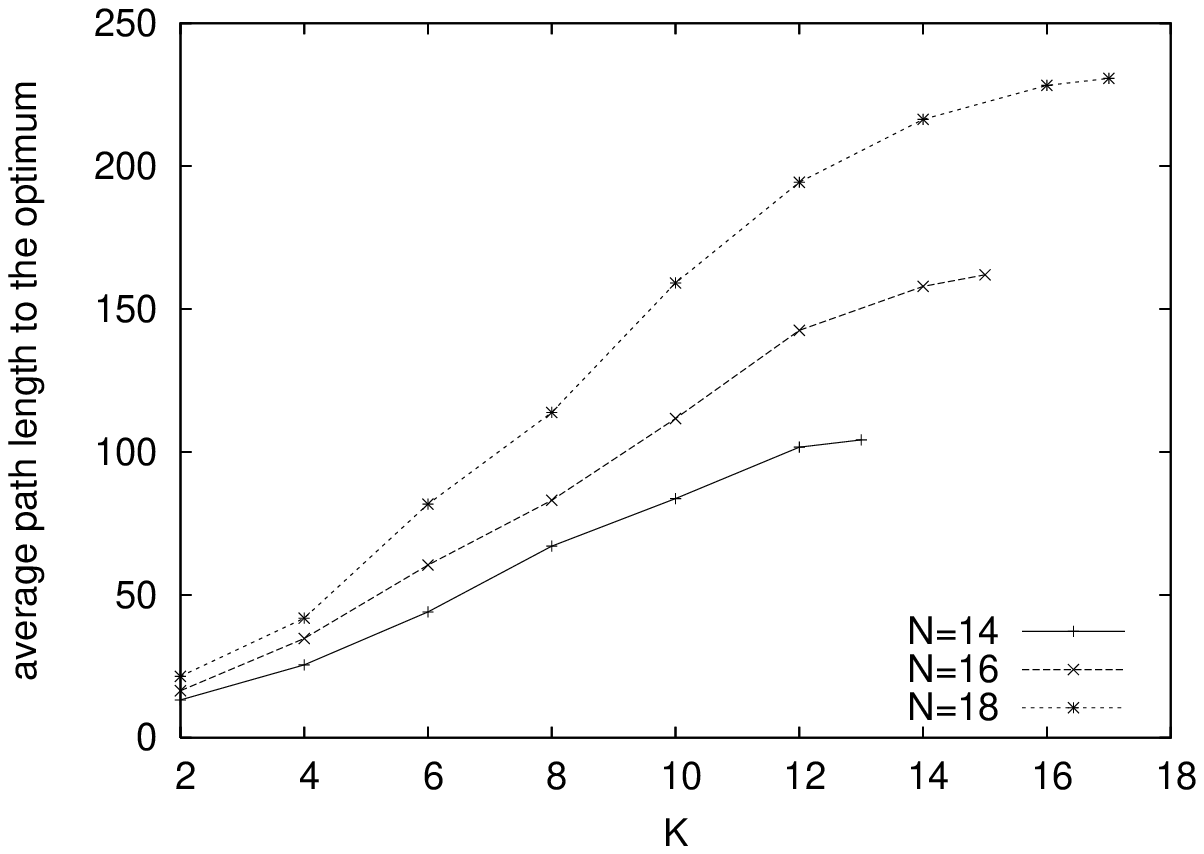} } \protect \\
    \vspace{-0.3cm}
\caption{Average distance (shortest path) between nodes (top), and
average path length to the optimum from all the other basins
(bottom). \label{fig:distances}}
\end{center}
\end{figure}

The average shortest path lengths $\bar d$ are listed in the sixth
column of table~\ref{tab:statistics}. Fig.~\ref{fig:distances} (top)
is a graphical illustration of the average shortest path length
between optima for all the studied $NK$ landscapes. Notice that the
shortest path increases with $N$, this is to be expected since the
number of optima increases exponentially with $N$. More
interestingly, for a given $N$ the shortest path increases with $K$,
up to $K = 10$, and then it stagnates and even decreases slightly
for the $N = 18$. This is consistent with the well known fact that
the search difficulty in $NK$ landscapes increases with $K$.
However, some paths are more relevant from the point of view of a
stochastic local search algorithm following a trajectory over the
maxima network. In order to better illustrate the relationship of
this network property with the search difficulty by heuristic local
search algorithms, Fig.~\ref{fig:distances} (bottom) shows the
shortest path length to the global optimum from all the other optima
in the landscape. The trend is clear, the path lengths to the
optimum increase steadily with increasing $K$.

\subsubsection{Weight Distribution}

\begin{figure}[!ht]
\begin{center}
\begin{tabular}{c}
{\tiny $N=16$}\\
\includegraphics[width=6.3cm]{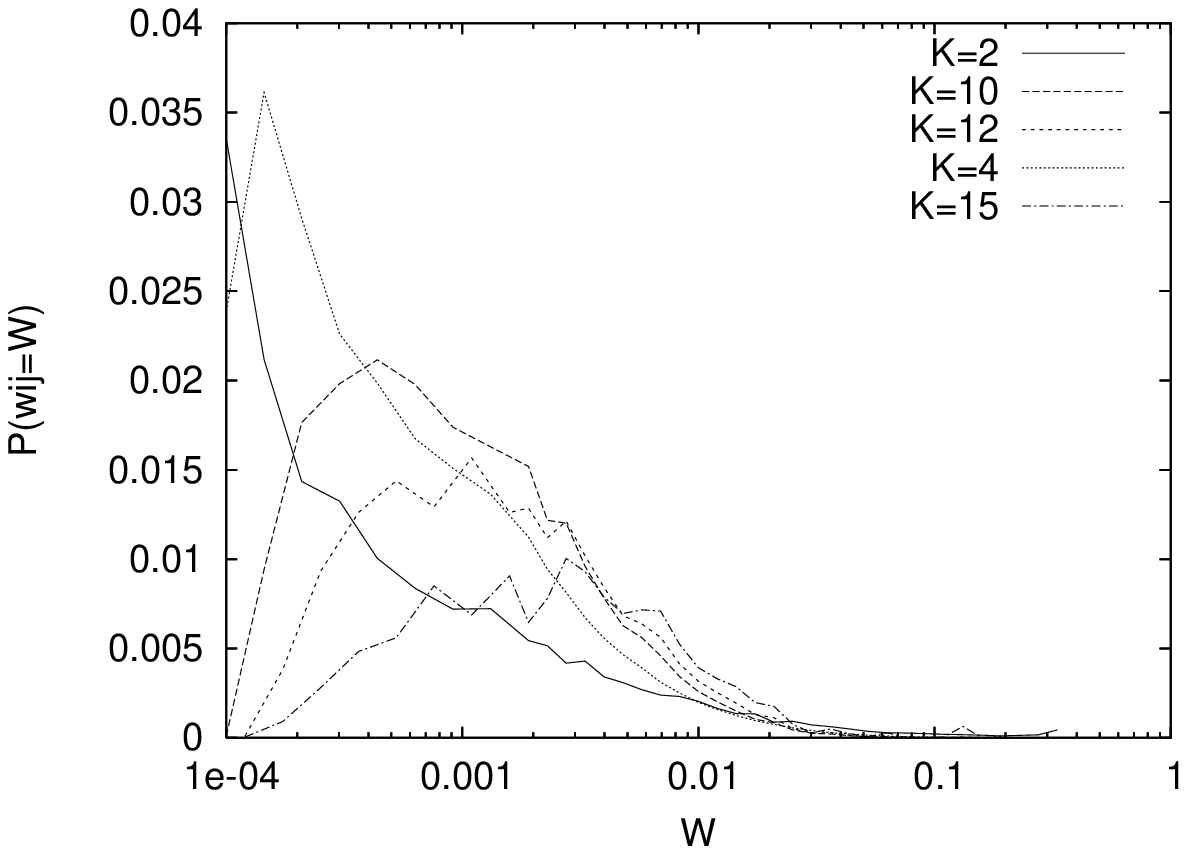}\\
{\tiny $N=18$}\\
\includegraphics[width=6.3cm]{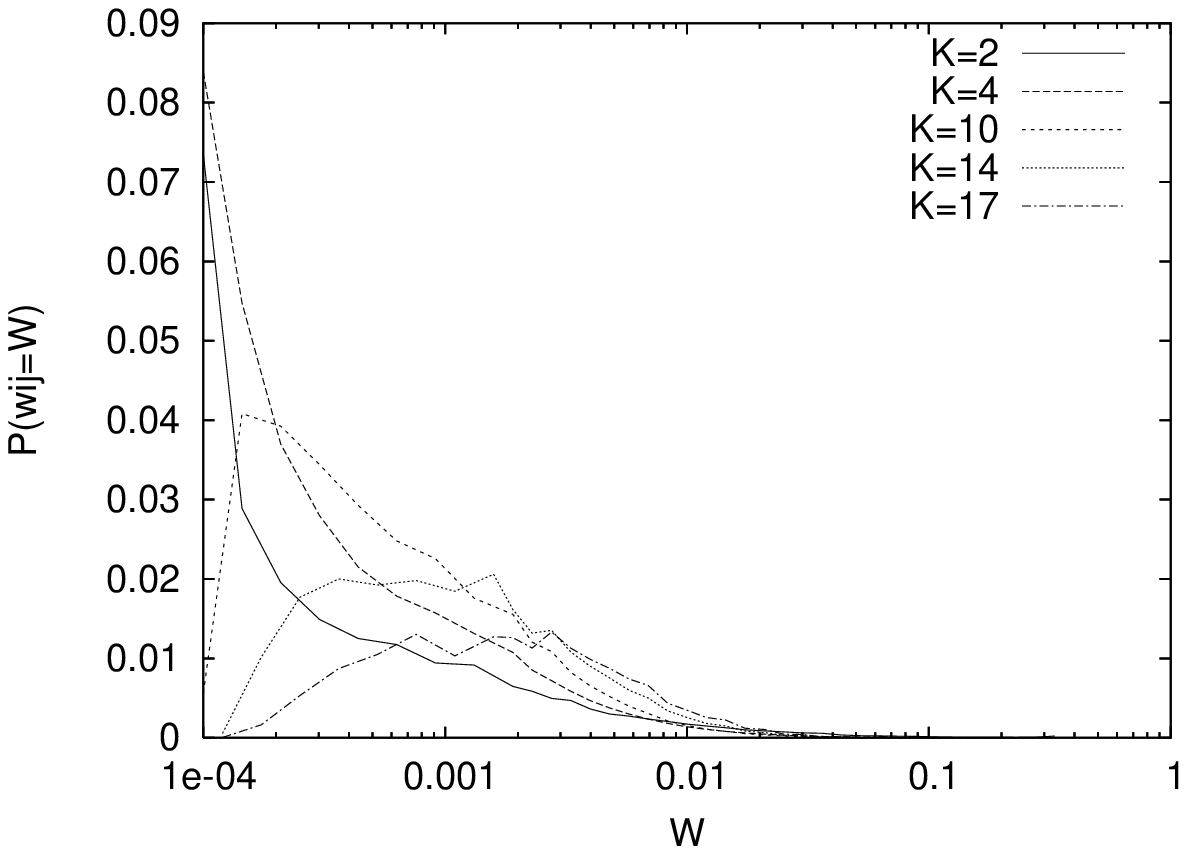} \\
\end{tabular}
\vspace{-0.3cm} \caption{Probability distribution of the network
weights $w_{ij}$ with $j\not=i$ in logscale on x-axis. Averages of
30 instances for each $N$ and $K$ are
reported.\label{fig-Distri-Wij}}
\end{center}
\end{figure}

Here we report on the weight distributions $p(w)$ of the maxima
network edges. Fig.~\ref{fig-Distri-Wij} shows the empirical
probability distribution function for the cases $N=16$ and $N=18$
(logarithmic binning has been used on the x-axis). The case $N=14$
is similar but is not reported here because it is much more noisy
for $K=2$ and $4$ due to the small size of the graphs in these cases
(see table~\ref{tab:statistics}).

One can see that the weights, i.e. the transition probabilities
between neighboring basins are small. The distributions are far from
uniform and, for both $N=16$ and $N=18$, the low $K$ have longer
tails. For high $K$ the decay is faster. This seems to indicate
that, on average, the transition probabilities are higher for low
$K$.

\subsubsection{Disparity}

\begin{figure} [!ht]
\begin{center}
\begin{tabular}{c}
     {\tiny $N=16$} \\
    \mbox{\includegraphics[width=6.3cm]{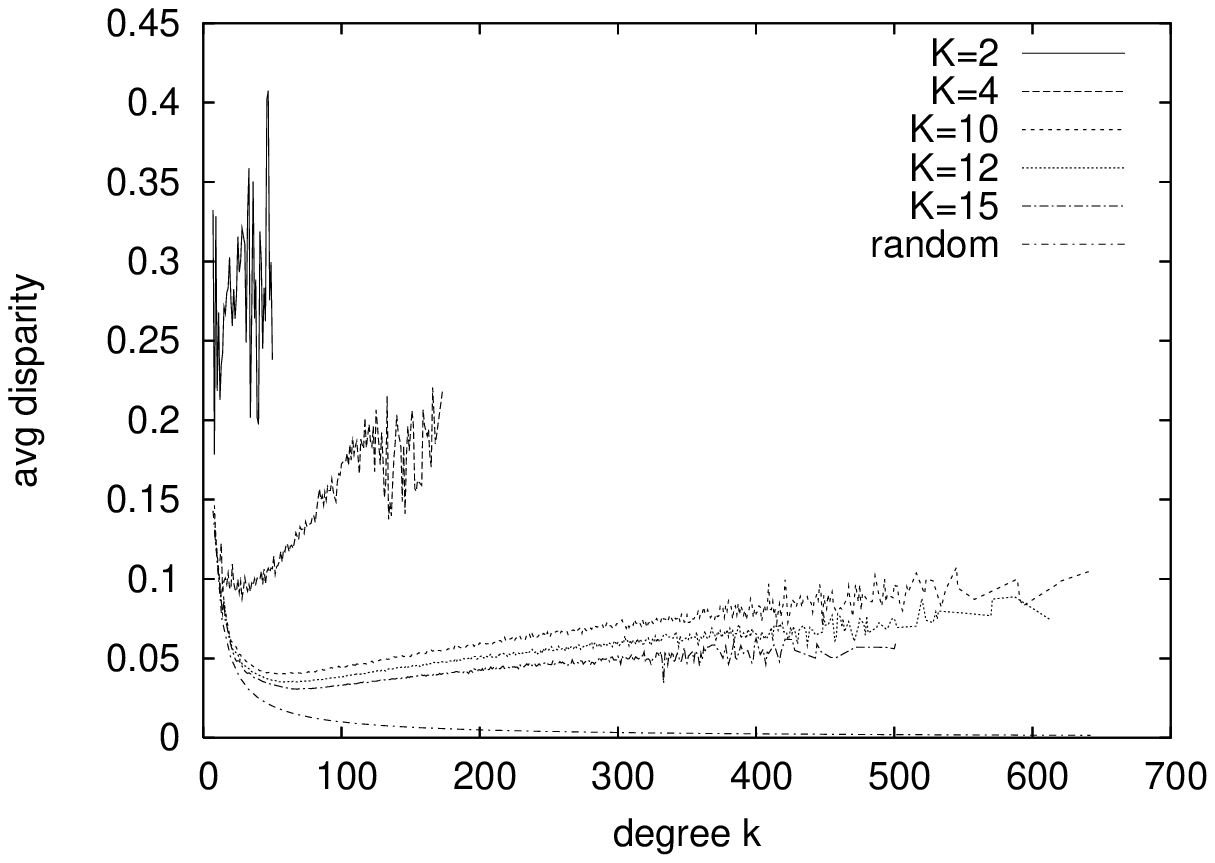}}\\
    {\tiny $N=18$} \\
    \mbox{\includegraphics[width=6.3cm]{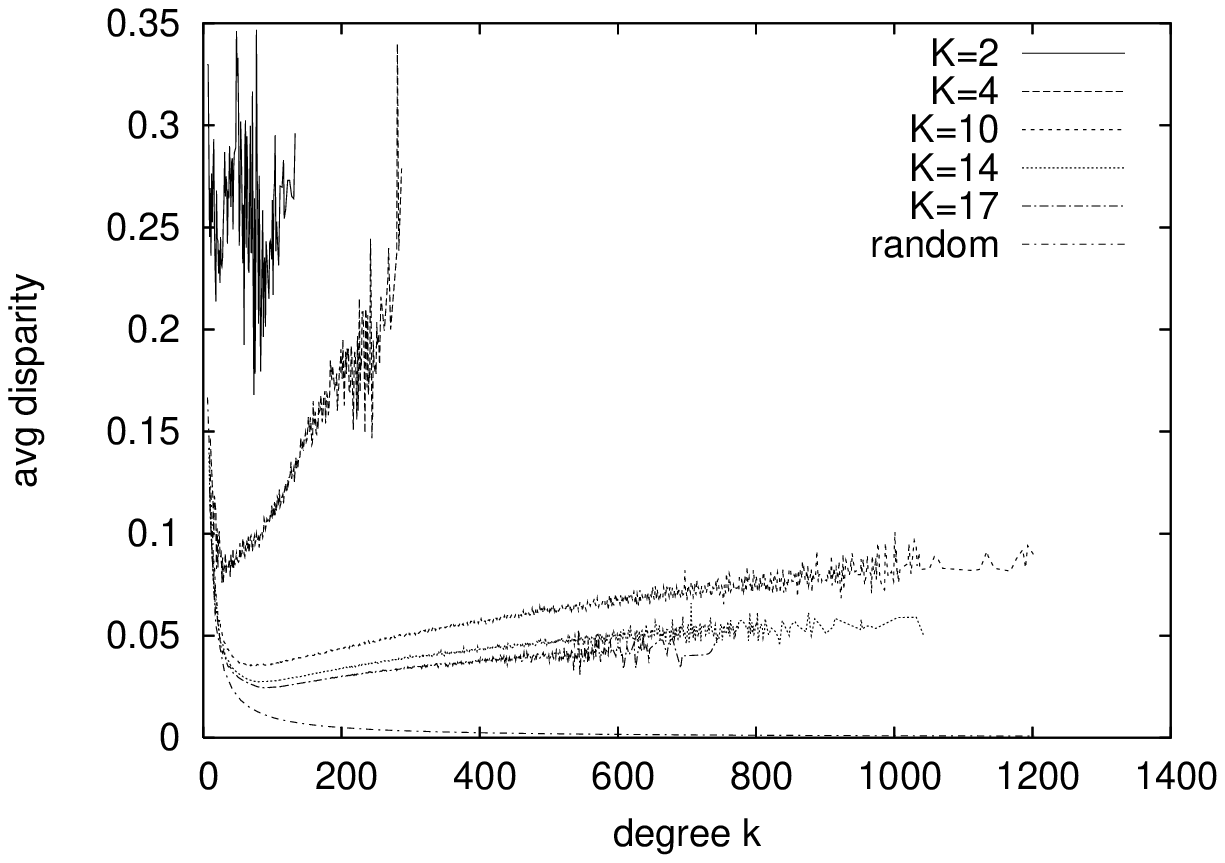}}\\
\end{tabular}
\vspace{-0.3cm} \caption{Average disparity,  $Y_{2}$, of nodes with
a given degree $k$. Average of $30$ independent instances for each
$N$ and $K$ are reported. The curve $1/k$ is also reported to
compare to random case.\label{fig-deg-disparity}}
\end{center}
\end{figure}

Fig.~\ref{fig-deg-disparity} depicts the disparity coefficient as
defined in the previous section for $N=16, 18$. An interesting
observation is that the disparity (i.e.~dishomogeneity) in the
weights of a node's outcoming links tends to decrease steadily with
increasing $K$. This reflects that for high $K$ the transitions to
other basins tend to become equally likely, which is another
indication that the landscape, and thus its representative maxima
network, becomes more random and difficult to search.

When $K$ increases, the number of edges increases and the number of
edges with a weight over a certain threshold increases too (see
fig.~\ref{fig-Distri-Wij}). Therefore, for small $K$, each node is
connected with a small number of nodes each with a relative high
weight. On the other hand,  for large $K$, the weights become more
homogeneous in the neighbourhood, that is, for each node, all the
neighboring basins are at similar distance.

If we suppose that edges with higher weights are likely to be
connected to nodes with larger basins (an intuition that we need to
confirm in future work). Then, as the larger basins tend to have
higher fitness (see Fig.~\ref{fig:cor_fit-size}), the path to higher
fitness values would be easier to find for lower $K$ than for larger
$K$.

\subsubsection{Boundary of basins.}

\begin{figure} [!ht]
\begin{center}
    \mbox{\includegraphics[width=6.3cm]{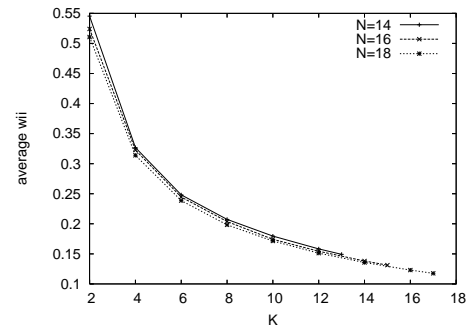} } \protect \\
\vspace{-0.3cm} \caption{Average weight $w_{ii}$ according to the
parameters $N$ and $K$. \label{fig:wii}}
\end{center}
\end{figure}

Fig. \ref{fig:wii} shows the averages, over all the nodes in the
network, of the weights $w_{ii}$ (i.e the probabilities of remaining
in the same basin after a bit-flip mutation). Notice that the
weights $w_{ii}$ are much higher when  compared to those $w_{ij}$
with $j \not= i$ (see  Fig.~\ref{fig-Distri-Wij}). In particular,
for $K=2$, $50\%$ of the random bit-flip mutations will produce a
solution within the same basin of attraction. These average
probabilities of remaining within the same basin, are  above $12 \%$
for the higher values of $K$. Notice that the averages are nearly
the same regardless the value of $N$, but decrease with the
epistatic parameter $K$.

The exploration of new basins with the random bit-flip mutation
seems to be, therefore, easier for large $K$ than for low $K$. But,
as the number of basins increases, and the fitness correlation
between neighboring solutions decreases with increasing $K$, it
becomes harder to find the global maxima for large $K$. This result
suggests that the dynamic of stochastic local search algorithms on
$NK$ landscapes with large $K$ is different than that with lower
values of $K$, with the former engaging in more random exploration
of basins.

The boundary of a basin of attraction can be defined as the set of
configurations within a basin that have at least one neighbor's
solution in another basin. Conversely, the interior of a basin is
composed by the configurations that have all their neighbors in the
same basin. Table~\ref{tab:inner} gives the average number of
configurations in the interior of basins (this statistic is computed
on $30$ independent landscapes). Notice that the size of the basins'
interior is below $1\%$ (except for $N=14$, $K=2$). Surprisingly,
the size of the basins' boundaries is nearly the same as the size of
the basins themselves. Therefore, the probability of having a
neighboring solution in the same basin is high, but nearly all the
solutions have a neighbor solution in another basin. Thus, the
interior basins seem to be \textit{"hollow"}, a picture which is
far from the smooth standard  representation of landscapes in 2D
with real variables where the basins of attraction are visualized as
real mountains.

\enlargethispage{\baselineskip}

\begin{table}
\vspace{-0.3cm} \caption{Average (on $30$ independent landscapes for
each $N$ and $K$) of the mean sizes of the basins
interiors.\label{tab:inner}}
\begin{center}
\begin{tabular}{|l|l|l|l|}
\hline
K & $N=14$ & $N=16$ & $N=18$ \\
\hline
2   & $0.0167_{0.02478}$ & $0.0050_{0.00798}$ & $0.0028_{0.00435}$ \\
4   & $0.0025_{0.00065}$ & $0.0012_{0.00025}$ & $0.0006_{0.00010}$ \\
6   & $0.0029_{0.00037}$ & $0.0014_{0.00015}$ & $0.0007_{0.00009}$ \\
8   & $0.0043_{0.00045}$ & $0.0022_{0.00012}$ & $0.0011_{0.00006}$ \\
10  & $0.0055_{0.00041}$ & $0.0031_{0.00015}$ & $0.0018_{0.00006}$ \\
12  & $0.0061_{0.00041}$ & $0.0040_{0.00014}$ & $0.0025_{0.00007}$ \\
13  & $0.0059_{0.00029}$ &                   &                   \\
14  &                   & $0.0045_{0.00012}$ & $0.0031_{0.00004}$ \\
15  &                   & $0.0044_{0.00014}$ &                   \\
16  &                   &                   & $0.0035_{0.00005}$ \\
17  &                   &                   & $0.0034_{0.00006}$ \\
\hline
\end{tabular}
\end{center}
\end{table}
\vspace{-0.3cm}

\section{Conclusions}
\label{conclusions}

We have proposed a new characterization of combinatorial fitness
landscapes using the family of $NK$ landscapes as an example. We
have used an extension of the concept of inherent networks proposed
for energy surfaces~\cite{doye02} in order to abstract and simplify
the landscape description. In our case the inherent network is the
graph where the nodes are all the local maxima and the edges
accounts for transition probabilities (using the bit-flip operator)
between the local maxima basins of attraction. We have exhaustively
obtained these graphs for $N=\{14, 16, 18\}$, and for all even
values of $K$, plus $K=N-1$, and conducted a network analysis on
them. Our guiding motivation has been to relate the statistical
properties of these networks, to the search difficulty of the
underlying combinatorial landscapes when using stochastic local
search algorithms (based on the bit-flip operator) to optimize them.
We have found clear indications of such
relationships, in particular:\\
The clustering coefficients suggest that, for high values of $K$,
the transition between a given pair of neighboring basins is less
likely to occur.\\
The shortest paths increase with $N$ and, for a given $N$, they clearly increase with
higher $K$.\\
The weight distributions indicate that, on average, the
transition probabilities are higher for low $K$.\\
The disparity coefficients reflect that for high $K$ the transitions to
other basins tend to become equally likely, which is an indication
of the randomness of the landscape.\\
The construction of the maxima networks requires the determination
of the basins of attraction of the corresponding landscapes. We
have thus also described the nature of the basins, and found  that
the size of the basin corresponding to the global maximum becomes
smaller with increasing $K$. The distribution of the basin sizes is
approximately exponential for all $N$ and $K$, but the basin sizes
are larger for low $K$, another indirect indication of the
increasing randomness and difficulty of the landscapes when $K$
becomes large. Furthermore, there is a strong positive correlation
between the basin size of maxima and their degrees.
Finally, we found that the size of the basins boundaries is
roughly the same as the size of basins themselves. Therefore, nearly
all the configurations in a given basin have a neighbor solution in
another basin. This observation suggests a different landscape
picture than the smooth standard representation of 2D landscapes
where the basins of attraction are visualized as hilltops.

This study is our first attempt towards a topological and
statistical characterization of combinatorial
landscapes, from the point of view of complex networks analysis.
Much remains to be done. The results  should be
confirmed for larger instances of $NK$ landscapes. This will require
good sampling techniques, or theoretical studies since exhaustive
sampling becomes quickly impractical. Other landscape types should
also be examined, such as those containing neutrality, which are
very common in real-world applications. Finally, the landscape
statistical characterization is only a step towards implementing
good methods for searching it. We thus hope that our results will
help in designing or estimating efficient search techniques and
operators.

\enlargethispage{\baselineskip}

\paragraph{Acknowledgements.}
Thanks to Professor Edmund Burke for useful advice and support. This
work was partially funded by EPSRC grant number EP/D061571/1.

\footnotesize
\bibliographystyle{apalike}

\end{document}